\documentclass[letterpaper, 10pt, conference]{ieeeconf}
\IEEEoverridecommandlockouts 

\usepackage{tabularx}
\usepackage[utf8]{inputenc}
\usepackage{amsmath,amssymb,amsfonts}
\usepackage{textgreek}
\usepackage{graphicx}
\usepackage{color,bm}
\usepackage{wasysym}
\usepackage{tikz}
\usepackage{cite}
\usepackage{array,multirow}
\usepackage{float}
\usepackage{balance}
\usepackage[hidelinks]{hyperref}
\usepackage{mathtools, nccmath}
\DeclareGraphicsExtensions{.tif}
\usepackage[linesnumbered,ruled,vlined]{algorithm2e}
\usepackage{amsmath} 
\usepackage{algorithmic,algorithm2e}
\usepackage{booktabs}
\usepackage{adjustbox}
\usepackage{tabularx}
\usepackage{makecell}
\usepackage{cellspace}

\usepackage{geometry}
\usepackage{url}

\newcount\Comments  
\newcount\Revision 
\definecolor{darkblue}{rgb}{0,0,0.75}
\newcommand{\kibitzrev}[2]{\ifnum\Revision=0\textcolor{#1}{#2}\else\textcolor{black}{#2}\fi}

\title{\LARGE \textbf{Interaction-Aware Personalized Vehicle Trajectory Prediction Using Temporal Graph Neural Networks}}

\author{Amr Abdelraouf,
        Rohit Gupta,
        Kyungtae Han
\thanks{A. Abdelraouf, R. Gupta, and K. Han are with InfoTech Labs, Toyota Motor North America R\&D, 465 Bernardo Avenue, Mountain View, CA 94043 (e-mail:
amr.abdelraouf@toyota.com;
rohit.gupta@toyota.com;
kyungtae.han@toyota.com).}
}

\begin{document}
\maketitle

\begin{abstract}
Accurate prediction of vehicle trajectories is vital for advanced driver assistance systems and autonomous vehicles. Existing methods mainly rely on generic trajectory predictions derived from large datasets, overlooking the personalized driving patterns of individual drivers. To address this gap, we propose an approach for interaction-aware personalized vehicle trajectory prediction that incorporates temporal graph neural networks. Our method utilizes Graph Convolution Networks (GCN) and Long Short-Term Memory (LSTM) to model the spatio-temporal interactions between target vehicles and their surrounding traffic. To personalize the predictions, we establish a pipeline that leverages transfer learning: the model is initially pre-trained on a large-scale trajectory dataset and then fine-tuned for each driver using their specific driving data. We employ human-in-the-loop simulation to collect personalized naturalistic driving trajectories and corresponding surrounding vehicle trajectories. Experimental results demonstrate the superior performance of our personalized GCN-LSTM model, particularly for longer prediction horizons, compared to its generic counterpart. Moreover, the personalized model outperforms individual models created without pre-training, emphasizing the significance of pre-training on a large dataset to avoid overfitting. By incorporating personalization, our approach enhances trajectory prediction accuracy.
\end{abstract}
 
\section{Introduction}
The advent of vehicle-generated big data has sparked considerable interest in data-driven personalized advanced driver assistance systems (ADAS) \cite{hasenjager2019survey}. By leveraging the wealth of personalized driving patterns and insights expressed by drivers, personalization has the potential to enhance the performance of ADAS systems greatly. This, in turn, leads to improved driving experiences, increased driver acceptance, and greater utilization of ADAS functionalities. In recent years, many personalized ADAS applications have been proposed, including Adaptive Cruise Control \cite{zhao2022personalized}, Forward Collision Warning \cite{wang2015forward}, Lane Keeping Assistance\cite{wang2018learning}, among many others. Furthermore, personalization extends to battery electric vehicle (BEV) ADAS applications. Personalized range estimation, in particular, holds significant promise in mitigating range anxiety, as it enables more accurate predictions of available driving range for individual drivers' given their unique driving styles \cite{shen2023personalized}.

Vehicle trajectory prediction is a foundational component in many ADAS applications and autonomous vehicle (AV) systems \cite{abdelraouf2022trajectory}. It plays a significant role in various safety-critical ADAS applications, such as collision warning \cite{lyu2020vehicle}, automated braking \cite{zhang2021automated, wang2023towards}, and lane change prediction \cite{liao2022online}. The accuracy of trajectory prediction is crucial for enhancing the effectiveness of these safety-critical applications \cite{mahmoud2023impact}, enabling them to provide timely warnings while minimizing false positives. Moreover, in the context of autonomous vehicles, trajectory prediction plays a vital role in ensuring safe motion planning and overall system safety \cite{huang2022survey}.  Anticipating and responding to dynamic traffic situations helps to reduce the risk of accidents and instills trust in AV systems. Additionally, trajectory prediction can significantly contribute to Vehicle-to-Vehicle (V2V) communication, particularly in intent-sharing \cite{wang2022conflict} and negotiation applications \cite{correa2021impact}. By facilitating vehicular communication of their intentions to others, trajectory prediction for intent-sharing promotes cooperative behavior in mixed traffic, improves traffic flow, and enhances overall safety.

Recent advancements in trajectory prediction have emphasized the significance of vehicle interactions \cite{huang2022survey}. Interaction-aware trajectory prediction considers the spatio-temporal relationship between the target vehicle and its surrounding neighbors. In the recent past, some of the most effective approaches for short-term trajectory prediction combined graph neural networks (GNNs) to model the spatial domain and recurrent neural networks (RNNs) to capture the temporal domain for vehicle interactions \cite{li2019grip++, zhao2020gisnet}.

Despite the substantial body of research dedicated to vehicle trajectory prediction, a notable research gap persists in personalized vehicle trajectory prediction. Specifically, there is limited exploration of interaction-aware personalized trajectory prediction methods \cite{ma2023cemformer}. A major challenge stems from the lack of individual driver trajectory data which captures the driver's trajectory in addition to the driver's surrounding vehicle trajectories for an extended period of time.

This paper presents a novel approach for generating personalized interaction-aware vehicle trajectory predictions. Our method builds upon state-of-the-art techniques and combines graph convolution networks (GCN) with Long Short-Term Memory (LSTM) models to effectively capture the dynamics of vehicle interactions. To obtain individual driver trajectories which can support interaction-aware modeling, we utilized human-in-the-loop simulation using the CARLA driving simulator \cite{dosovitskiy2017carla}. The proposed personalization framework leverages transfer learning. Initially, a base model was pre-trained on a large-scale vehicle trajectory dataset named CitySim \cite{zheng2022citysim}. Subsequently, the model was fine-tuned for each driver using their individual driving data, resulting in a unique model for each driver tailored to their specific driving characteristics. Experiments indicated that our proposed personalized approach generates more accurate future trajectories compared to generic trajectory prediction. The proposed method can be used to personalize various ADAS applications, such as collision warning, enhancing their accuracy while reducing their false positive rate. The improvements can contribute to a better driving experience by improving safety, reliability, and driver trust.

In summary, the contribution of this paper is three-fold:
\begin{enumerate}
\item We introduced a new method for predicting future vehicle trajectories that incorporates personalized driving characteristics. Our approach employed transfer learning to first learn generic interaction-aware driving patterns, which were then fine-tuned using driver-specific trajectories.
\item To obtain driver-specific trajectories which can support interaction-aware trajectory prediction, we conducted a human-in-the-loop simulation experiment that captured naturalistic driving behavior and surrounding vehicle trajectories.
\item We demonstrated the effectiveness of our approach through experiments and analysis, which showed that our interaction-aware personalized trajectory prediction outperforms generic trajectory prediction methods.
\end{enumerate}

The remainder of this paper is structured as follows: Section II provides an overview of related work, Section III describes the personalization scheme and model architecture, Section IV describes the personalized driving data and large-scale trajectory dataset used for experimentation, Section V presents the experimental results, and Section VI concludes the paper and discusses future work.

\section{Related Work}

\subsection{Vehicle Trajectory Prediction}
In recent years, there have been several comprehensive surveys focusing on vehicle trajectory prediction \cite{lefevre2014survey, mozaffari2020deep, huang2022survey}. Notably, Huang et al. \cite{huang2022survey} conducted a recent survey that examined various trajectory prediction methodologies. These approaches were categorized into physics-based methods, classic machine learning methods, reinforcement learning-based methods, and deep learning-based methods. Physics-based models are often regarded as the simplest approach to trajectory prediction. While these methods may be suitable for short prediction horizons, their accuracy deteriorates as the output horizon increases. Physics-based methods encompass a range of models, from basic ones like constant velocity and constant acceleration to more sophisticated techniques like Kalman Filtering (KF) methods \cite{ammoun2009real}. Classic machine learning methods utilize a data-driven approach for trajectory prediction. These algorithms incorporate various techniques such as Gaussian Process (GP) \cite{tran2014online, vasquez2004motion}, Support Vector Machine (SVM) \cite{aoude2010threat, kumar2013learning}, and Hidden Markov Models (HMMs) \cite{deo2018would, zhang2020research}. Reinforcement learning approaches leverage expert demonstrations to learn optimal driving policies by maximizing the expected reward. Example algorithms include Inverse Reinforcement Learning (IRL) \cite{xu2020learning} and Generative Adversarial Imitation Learning (GAIL) \cite{kuefler2017imitating}.

In the recent past, deep learning-based approaches have demonstrated state-of-the-art performance for vehicle trajectory prediction, particularly given the increased availability of large trajectory datasets. The superior performance can be attributed to the innate effectiveness of deep learning at capturing the intricate patterns present in large datasets \cite{mozaffari2020deep, abdelraouf2021utilizing}. Previous research has focused on utilizing sequential deep neural networks to address the temporal nature of the vehicle trajectory prediction problem. Methods such as Recurrent Neural Networks (RNNs) \cite{xin2018intention, park2018sequence} and sequence-to-sequence Transformers \cite{giuliari2021transformer} have been employed for this purpose.

Vehicle interactions play a significant role in vehicle trajectory prediction. To effectively incorporate these interactions within the deep learning modeling framework, prior research has frequently integrated a sequential neural network module with a spatial counterpart to create an end-to-end spatio-temporal trajectory prediction network. For instance, Convolutional Neural Networks (CNNs) have often been utilized to model the spatial context for the trajectory prediction problem \cite{deo2018convolutional, chandra2019traphic, xie2020motion}. Deo et al. \cite{deo2018convolutional} employed CNNs and LSTM to create a convolutional social pooling algorithm. The authors utilized a CNN-LSTM architecture to capture the trajectories of neighboring vehicles, extracting embeddings that were then utilized to generate multi-modal trajectories based on vehicle maneuvers. In recent publications, Graph Neural Networks (GNNs) have gained popularity for modeling the spatial domain of vehicle interactions \cite{li2019grip++, zhao2020gisnet, zeng2021lanercnn, su2020graph}. GNNs are well-suited for this task as they effectively capture the relationships among vehicles in the neighborhood. Each vehicle is represented as a graph node, and the interactions between vehicles are represented by edges in the graph.

\begin{figure*}[t!]
    \centering
    \includegraphics[width=\linewidth]{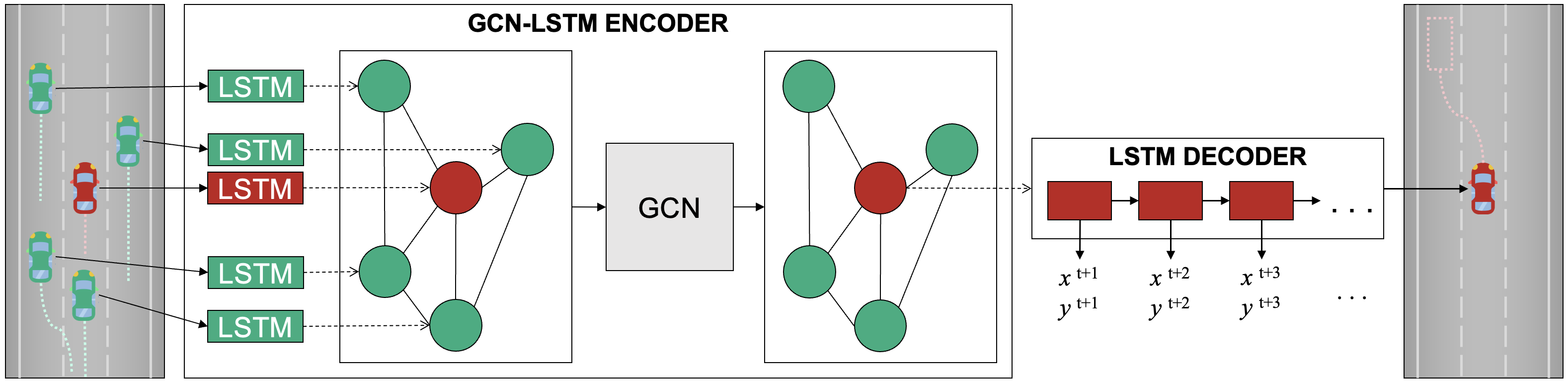}
    \caption{Proposed trajectory prediction network architecture}
    \label{fig:architecture}
\end{figure*}

\subsection{Personalized Trajectory Prediction}
A few research efforts proposed methods for personalized trajectory prediction. However, due to lack of long-term interaction-aware personal trajectory data, many of these researches proposed to first group driving behavior into distinct clusters \cite{xing2019personalized, yi2018trajectory}. For example, Xing et al. \cite{xing2019personalized} introduced an approach for trajectory prediction that starts by clustering the driving style of the target vehicle into one of three categories: conservative, moderate, or aggressive using Gaussian Mixture Models (GMM). The method utilized a pre-trained LSTM network and subsequently fine-tuned a regression head tailored to each of the three driving styles.

Other approaches aimed to learn individual driver profiles by constructing personalized reward functions using Inverse Reinforcement Learning (IRL) \cite{huang2021driving, liao2023driver}. However, the approach in \cite{huang2021driving} was developed without leveraging long-term personalized trajectories and was tested on a few seconds of individual driving periods. Moreover, the approach presented in \cite{liao2023driver} did not consider the interactions between the target vehicle and its neighboring vehicles.

This paper aimed to address the literature gap in personalized trajectory prediction by first collecting extended personalized vehicle trajectory. To facilitate interaction-aware trajectory prediction, the dataset captured target vehicles' surrounding traffic trajectory data as well. Next, an interaction-aware approach which utilized temporal graph neural networks was introduced. Transfer learning techniques were incorporated to create personalized models for each driver. Finally, the effectiveness of the approach was evaluated using the collected individual driving data.

\section{Methodology}

The proposed model is illustrated in Figure \ref{fig:architecture}. It presents an architecture designed to predict the future trajectory of the target vehicle by processing a traffic graph centered around it. The model follows an encoder-decoder structure, with a GCN-LSTM network serving as the encoder and an LSTM network serving as the decoder.

\subsection{Traffic Graph}
The input graph is represented as an undirected graph \( G = (V, E) \), where the set of nodes \( v_i \in V \) of size \( N \) consists of the target vehicle and the vehicles in its neighborhood. The set of edges \( (v_i, v_j) \in E \) represents the connections between vehicles in close proximity, satisfying two conditions: they are within 1 lane of each other and their distance is less than or equal to a threshold \( \tau \).

Each vehicle \( v_i \) is associated with a node feature vector \( X_i \in \mathbb{R}^{T_{\text{in}}+1 \times F} \), containing its historical trajectory for the past \( T_{\text{in}} \) time steps in addition to the current time step. Each element within a time step contains the vehicle's \( x \)-coordinate, \( y \)-coordinate, and speed. The historical trajectory \( X_i \) of a vehicle \(v_{i}\) is defined as:

\begin{equation}
X_i = \{ C^{t - T_{in}}_i, \ldots, C^{t-1}_i, C^{t}_i \}
\end{equation}

where each element \( C^{t}_i \) is defined as:

\begin{equation}
C^{t}_i = \{ x^t_i, y^t_i, s^t_i \}
\end{equation}

\subsection{Interaction Modeling}

The proposed encoder model combined an LSTM network for temporal feature embedding and a GCN network for capturing the spatial interaction between neighboring vehicles. The LSTM network was specifically designed to capture sequential dependencies, enabling it to extract valuable features from each vehicle's historical trajectory data. By sharing weights across LSTMs, the model ensured consistent representations. The LSTM generated 1-dimensional embeddings \(X^{embed}\), which served as node feature inputs for the GCN layers.

Graph Convolution Networks, initially introduced in \cite{kipf2016semi}, were specifically designed as an alternative to CNNs for effective modeling of graph-based structures. GCNs require two inputs: an adjacency matrix and a set of features. In our implementation, the adjacency matrix \( A \in \mathbb{R}^{N \times N} \) represents the connectivity between nodes in the graph, where \( N \) corresponds to the number of nodes. \( A_{i,j} = A_{j,i} = 1 \) if there exists an edge \( (v_i, v_j) \), and 0 otherwise. The set of features \( X \in \mathbb{R}^{N \times D} \) represents the input feature vector for each node in the graph, where \( D \) denotes the number of features. Notably, in our implementation, \( X \) is constructed from the output of the LSTM embeddings \( X^{embed} \) and \( D \) is equal to the size of the embedding output from the encoder LSTM.

Building upon the methodology described in \cite{zhao2020gisnet}, our approach incorporates a 2-layer GCN module, which performs the following operation:

\begin{equation}
\text{GCN}(X^{embed},A) = \hat{D}^{-\frac{1}{2}}\hat{A}\hat{D}^{-\frac{1}{2}}H^{(0)}W^{(1)}
\end{equation}

Here, \(H^{(0)}\) is computed as:

\begin{equation}
H^{(0)} = \text{RELU}(\hat{D}^{-\frac{1}{2}}\hat{A}\hat{D}^{-\frac{1}{2}}X^{embed}W^{(0)})
\end{equation}

In the equations above, a node-wise self-connection was established by adding the input adjacency matrix \(A\) to the identity matrix \(I\), denoted as \(\hat{A} = A + I\). This self-connection allows vehicle nodes to incorporate their own historical trajectories during GCN propagation. Furthermore, the weights of the adjacency matrix were normalized by the node degree using the diagonal matrix \(\hat{D}_{i,i} = \sum_{j}^{N} A_{i,j}\). Finally, \(W^{(0)}\) and \(W^{(1)}\) represented the trainable parameters of the first and second GCN layers, respectively.

\subsection{Future Trajectory Generation}

The decoder layer of the proposed model is responsible for generating the future trajectory of the target vehicle. It begins by extracting the GCN embedding state of the target node and utilizing it as input for an LSTM module. The decoder LSTM captures the temporal dependencies between the output time steps. Subsequently, the model generates the output \( Y_i \in \mathbb{R}^{T_{\text{out}} \times 2} \), which contains the \(x\) and \(y\) coordinates of the future trajectory for each time step in the output horizon \(T_{\text{out}}\).

The future trajectory \(Y_i\) for a target vehicle \(v_i\) is defined as follows:

\begin{equation}
Y_i = \{ \{x^{t + 1}_i, y^{t + 1}_i\}, \ldots, \{x^{t + T_{\text{out}}}_i, y^{t + T_{\text{out}}}_i\} \}
\end{equation}

\subsection{Model Training}

The proposed model was trained using backpropagation, with the loss function $\mathcal{L}$ defined as:

\begin{equation}
\mathcal{L}(Y, \hat{Y}) = \frac{1}{T_{\text{out}} \cdot M} \sum_{i=1}^{M} \sum_{t=1}^{T_{\text{out}}} \left| x_i^{t} - \hat{x}_i^{t} \right| + \left| y_i^{t} - \hat{y}_i^{t} \right|
\end{equation}

In this equation, \( x_i^{t} \) and \( y_i^{t} \) represent the true \( x \)-coordinate and \( y \)-coordinate of example \( i \) at time step \( t \), respectively. The predicted \( x \)-coordinate and \( y \)-coordinate at time step \( t \) are denoted as \( \hat{x}_i^{t} \) and \( \hat{y}_i^{t} \), respectively. \(M\) is the number of examples used for training and \(T_{out}\) is the output prediction horizon.

\subsection{Personalization}

\begin{figure}[t!]
    \centering
    \includegraphics[width=\columnwidth]{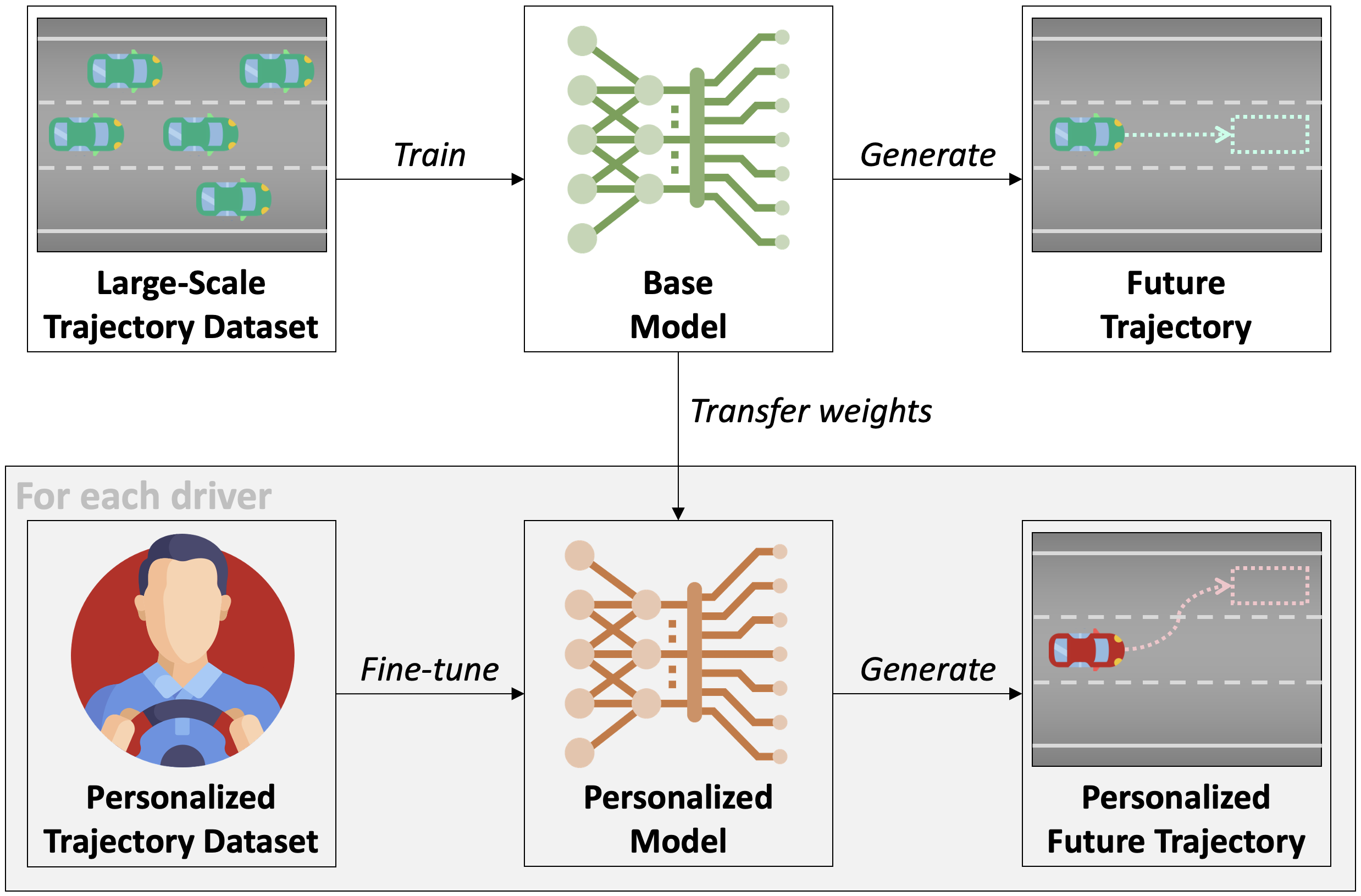}
    \caption{Transfer learning-based trajectory prediction personalization}
    \label{fig:personalization}
\end{figure}

The proposed personalization scheme, depicted in Figure \ref{fig:personalization}, involves two main steps. Initially, a base model is constructed by training the proposed architecture on a large-scale vehicle trajectory dataset. Subsequently, personalized models are created for each driver using transfer learning. This is achieved by loading the weights of the pre-trained base model and fine-tuning it using the driver's own trajectories. During fine-tuning, the encoder network weights remain frozen to preserve the learned spatio-temporal interaction embedding representations, while the decoder weights are updated. This approach allows for personalized trajectory prediction by leveraging the base model's learned features and adapting them to individual driver characteristics.

\section{Data Description}

\begin{table*}[t]
\centering
\caption{Trajectory prediction RMSE (m)}
\label{tab:results}
\setlength{\extrarowheight}{4pt}
\begin{tabularx}{0.8\linewidth}{
  |>{\centering\arraybackslash\bfseries\hsize=0.166\hsize}X
  |>{\centering\arraybackslash\hsize=0.166\hsize}X
  |>{\centering\arraybackslash\hsize=0.166\hsize}X
  |>{\centering\arraybackslash\hsize=0.166\hsize}X
  |>{\centering\arraybackslash\hsize=0.166\hsize}X
  |>{\centering\arraybackslash\hsize=0.166\hsize}X|}
\hline
& & & & \\[-11pt]
\textbf{\makecell{Prediction\\Horizon (s)}} & \textbf{CV} & \textbf{\makecell{Seq2seq\\LSTM}} & \textbf{\makecell{Generic\\GCN-LSTM}} & \textbf{\makecell{Individual\\GCN-LSTM}} & \textbf{\makecell{Personalized\\GCN-LSTM}} \\[2pt]
\hline
\textbf{1} & 1.427 & 1.168 & 0.834 & 1.063 & \textbf{0.763} \\
\textbf{2} & 2.360 & 1.546 & 1.344 & 1.533 & \textbf{1.191} \\
\textbf{3} & 3.573 & 2.741 & 1.820 & 1.965 & \textbf{1.529} \\
\textbf{4} & 5.439 & 3.275 & 2.547 & 2.670 & \textbf{2.022} \\
\textbf{5} & 6.109 & 5.885 & 5.351 & 4.486 & \textbf{3.024} \\[2pt]
\hline
\end{tabularx}
\end{table*}

\subsection{Personalized Trajectories}

To overcome the limited availability of long-term naturalistic driving data that can support interaction-aware trajectory prediction, we adopted a human-in-the-loop driver simulation approach. The CARLA driving simulator \cite{dosovitskiy2017carla} served as our platform, providing pre-implemented vehicle models, control and physics, road geometry, and traffic management modules. In our hardware setup, we utilized a G29 Logitech steering wheel, floor pedals, and a driving seat.

\begin{figure}[t!]
    \centering
    \includegraphics[width=\columnwidth]{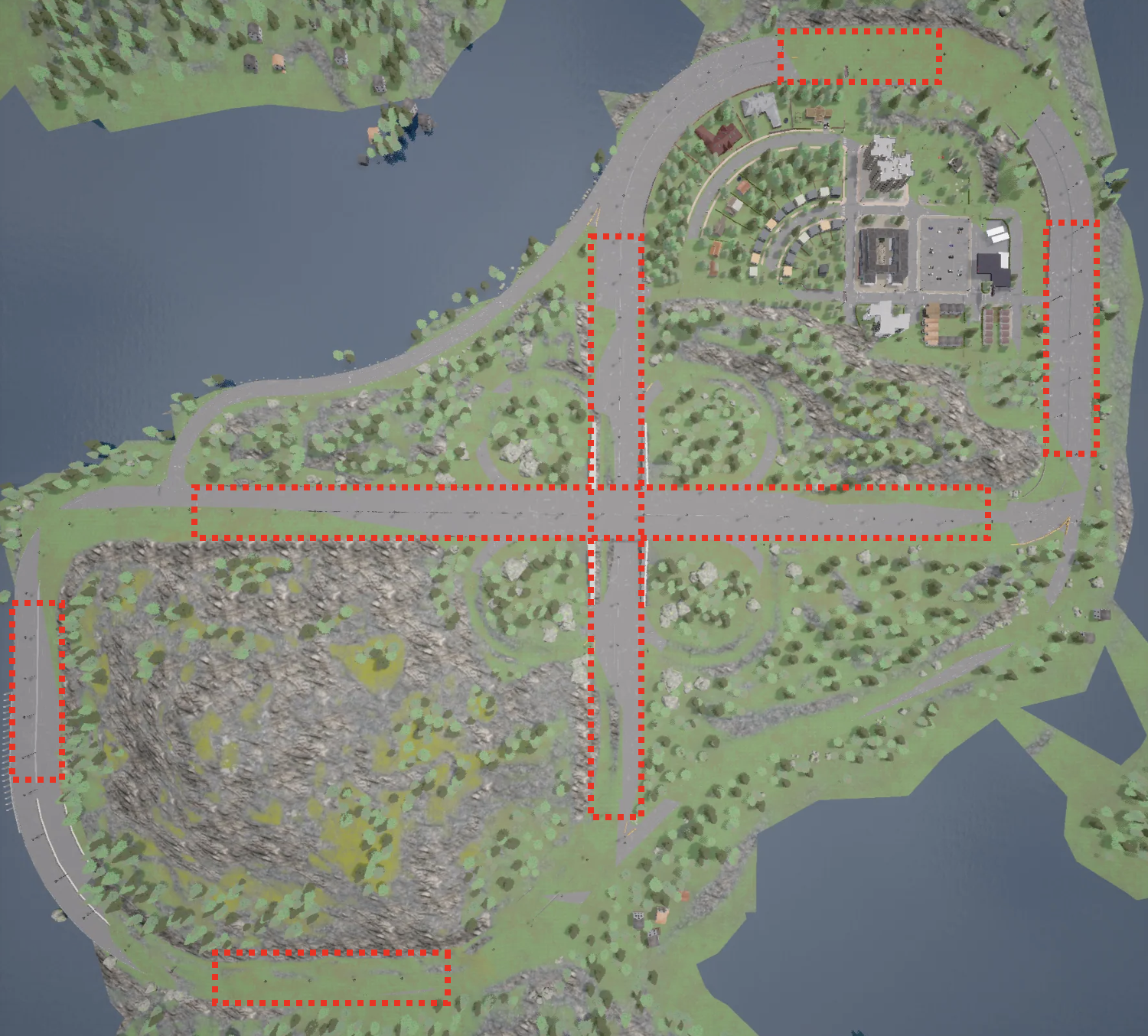}
    \caption{CARLA Town 04 map annotated with the portions of the road used for data collection}
    \label{fig:carla}
\end{figure}

Our experimentation focused on highway scenarios. Therefore, we utilized Town 04 in CARLA, which offers a continuous 3-lane highway configured in the shape of a figure-eight. We extracted the trajectories from the non-curved segments of the highway for use in our experiment as illustrated in Figure \ref{fig:carla}.

We gathered data from a total of five distinct drivers for our study. Each driver received instructions to adhere to traffic laws and drive in a typical manner as they would on public roads. To capture a comprehensive range of driving behaviors, each participant completed four driving rounds, with each round lasting 10 minutes. Thus, we obtained a total of 40 minutes of personalized naturalistic driving data per driver. To capture the range of driving behaviors under diverse traffic conditions, we systematically varied the number of vehicles in the background across the rounds. Specifically, we set the number of vehicles in the background traffic with three different settings: 100, 200, and 300 vehicles. This variation enabled us to capture how different drivers adapt and respond to different traffic densities in our personalized dataset.

\subsection{Large-Scale Trajectories}

The CitySim dataset \cite{zheng2022citysim}, specifically the Freeway B scenario, was used for base model pre-training. This dataset includes bird's-eye view vehicle trajectories obtained from drone videos. Multiple drones were flown concurrently to capture continuous trajectories along an extended highway segment. The Freeway B scenario is a long continuous stretch of a non-curved highway. The dataset covered a duration of 35 minutes and included trajectory data from a total of 7,307 vehicles in various traffic conditions. The average trajectory length was 45 seconds, with a standard deviation of 24 seconds. The dataset provides a diverse range of traffic conditions, allowing the base model to learn from a wide range of scenarios encountered by numerous drivers on the freeway.

\section{Experiments and Results}

\subsection{Experimental Setup}

In this experiment, 10 minutes from each driver's personalized trajectories were used for evaluation. To generate future trajectories, 5 seconds of historical trajectories were utilized to predict the subsequent 5 seconds of future trajectories for the target vehicle. The time step interval employed was 0.5 seconds. Thus, the proposed model parameters \( T_{in}=T_{out}=10 \). Furthermore, in the input traffic graph, the distance threshold for edge creation $\tau$ was set to 100 feet.

\subsection{Evaluation Metrics}

Similar to previous trajectory prediction works, we adopted Root Mean Square Error (RMSE) for evaluation. RMSE measures the difference between the ground truth and the predicted value at time step $t$ using the following equation:
\begin{equation}
\mathrm{RMSE}^t = \sqrt{\frac{1}{M}\sum_{i=1}^{M}(x_i^t - \hat{x}_i^t)^2 + (y_i^t - \hat{y}_i^t)^2}
\end{equation}

In this equation, $x_i^t$ and $y_i^t$ represent the true coordinates at time step $t$, and $\hat{x}_i^t$ and $\hat{y}_i^t$ represent the predicted coordinates at time step $t$. $M$ is the number of samples. We compare the RMSE values at output horizons of 1 second to 5 seconds using 1-second intervals. The RMSE values were reported in meters.

\subsection{Baseline Models}

Several baseline models were used to compare the performance of the proposed model. The results compare between the following models:

\begin{itemize}
    \item Constant Velocity (CV): A constant velocity Kalman filter was employed to predict the future trajectory. It assumes constant velocity motion.
    \item Sequence-to-Sequence (Seq2seq) LSTM: This approach utilizes an LSTM encoder and LSTM decoder network. It does not incorporate a graph module for spatial interaction modeling.
    \item Generic GCN-LSTM: The base GCN-LSTM model was employed without any fine-tuning for individual drivers.
    \item Individual GCN-LSTM: Separate GCN-LSTM models were trained for each individual driver using their respective data. No pre-training was conducted for these models. 
\end{itemize}

\subsection{Results}

Table \ref{tab:results} compares the results of the proposed model against the baseline models. It contains output RMSEs for different prediction horizons. The results indicate that, while the output of the CV model is relatively comparable for shorter prediction horizons, it rapidly deteriorates as the prediction horizon increases. The generic GCN-LSTM outperforms the CV model and the Seq2seq LSTM model, highlighting the effectiveness of the spatio-temporal interaction-aware modeling approach.

\begin{figure}[t!]
    \centering
    \includegraphics[width=\columnwidth]{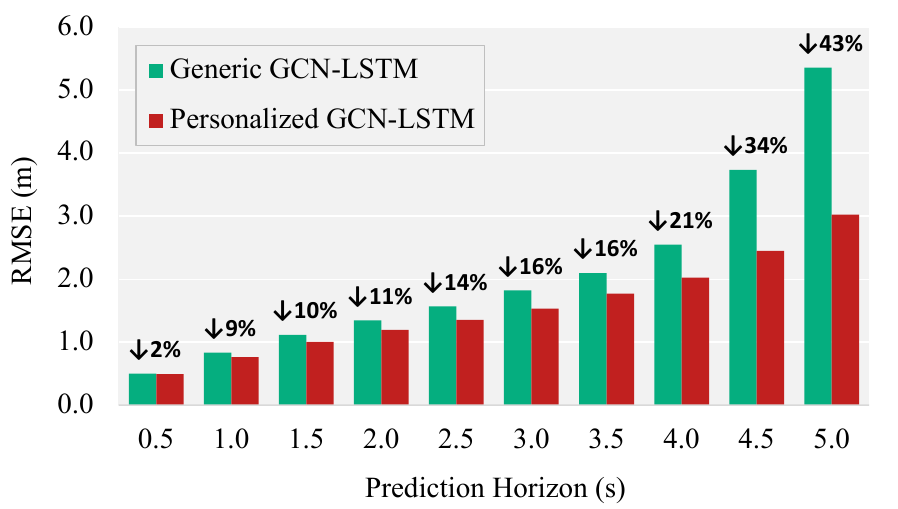}
    \caption{Percentage RMSE reduction of the personalized GCN-LSTM model compared to the generic GCN-LSTM model for different prediction horizons}
    \label{fig:graph}
\end{figure}

The results in Table \ref{tab:results} indicate that the personalized GCN-LSTM surpasses its generic counterpart, emphasizing the importance of personalization in vehicle trajectory prediction. Furthermore, the corresponding graph presented in Figure \ref{fig:graph} illustrates the RMSE reduction percentage achieved by the personalized GCN-LSTM model compared to the generic model. The results indicate that the further out the prediction horizon, the higher the error reduction margin. Notably, longer prediction horizons are more challenging to forecast given the higher level of uncertainty. Nevertheless, the results indicate that the personalized model improvement is more pronounced for longer horizons.

In order to assess the significance of the pre-training phase for the personalized GCN-LSTM model, its performance was compared to individual GCN-LSTM models trained exclusively on each driver's data. The findings presented in Table \ref{tab:results} demonstrate that the personalized GCN-LSTM model surpasses the individual model results. This observation implies that the individual model may encounter overfitting issues when solely trained on the driver's data. By leveraging a diverse range of driving scenarios during pre-training, the personalized model showcases superior generalization capabilities compared to the individual model.

\subsection{Impact of Personalization}

\begin{figure}[t!]
    \centering
    \includegraphics[width=\columnwidth]{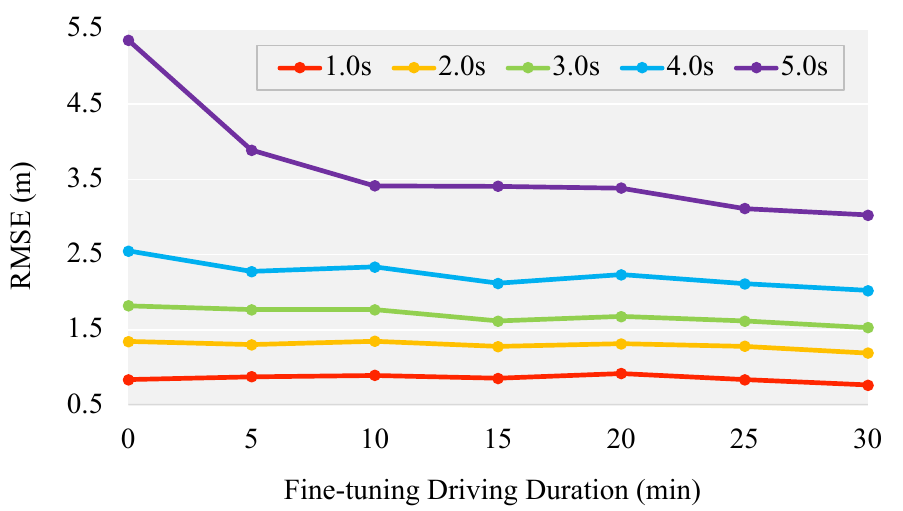}
    \caption{Impact of driving duration used for personalization on trajectory prediction RMSE}
    \label{fig:graph2}
\end{figure}

To further investigate the impact of personalization on vehicle trajectory prediction, the number of driving minutes used for fine-tuning was varied. As mentioned earlier, 40 minutes of personalized naturalistic driving data were collected for each driver, with 10 minutes allocated for testing. Subsequently, the number of driving minutes for model fine-tuning was varied from 5 to 30 minutes, and the RMSE was computed for each variation.

Figure \ref{fig:graph2} illustrates the relationship between the driving duration used for fine-tuning and the trajectory prediction RMSE for different horizons. The graph demonstrates a clear trend where an increase in the number of fine-tuning minutes leads to a decrease in RMSE. Furthermore, in alignment with the results depicted in Figure \ref{fig:graph}, it can be noticed that personalization has a more pronounced effect on longer prediction horizons. The graph trends suggest that a longer duration of personalized driving data positively impacts the trajectory prediction accuracy, particularly for longer prediction horizons.

\section{Conclusion and Future Work}

In this paper, we proposed a personalized interaction-aware trajectory prediction method that leverages GCNs and LSTMs to model the spatio-temporal vehicle interactions between the target vehicle and its neighboring vehicles. By employing transfer learning, we personalized the proposed model by pre-training it on a large-scale vehicle trajectory dataset and subsequently fine-tuning it for each unique driver, thus creating a personalized model that captures the unique patterns of individual drivers. To overcome the lack of availability of extended interaction-based personalized trajectories, we employed human-in-the-loop simulation to collect interaction-based personalized trajectories.

Through experiments, we demonstrated that the personalized GCN-LSTM model outperforms its generic counterpart, particularly for longer prediction horizons. This highlights the effectiveness of our personalized approach in capturing the driver-specific characteristics and improving trajectory predictions. Furthermore, our results indicated that the personalized model surpasses individual models created without pre-training, emphasizing the significance of pre-training on a large dataset to mitigate overfitting issues and enhance prediction accuracy.

The implications of personalized trajectory prediction are profound for various ADAS systems such as collision warning and lane keeping assist. By tailoring the predictions to individual patterns, the accuracy is improved, and the false positive rate is reduced. Consequently, these enhancements lead to greater safety and reliability within the systems, while instilling driver trust. Furthermore, personalized trajectory prediction can affect vehicle-to-vehicle (V2V) communication applications, specifically intent-sharing and road resource negotiations between vehicles. This is particularly relevant in mixed traffic scenarios, where both connected automated vehicles (CAVs) and manually driven connected vehicles interact.

Future work should focus on further validating the significance of personalization in interaction-aware trajectory prediction by collecting a larger dataset of personalized trajectories. In addition, although simulation data offers a convenient way to demonstrate the effectiveness of personalization, it is crucial to address the domain shift between simulation and real driving. Integrating real-world data into the training process will enhance the model's ability to generalize and accurately predict trajectories in practical driving scenarios.

\section*{Acknowledgment}
The contents of this work only reflect the views of the authors, who are responsible for the facts and the accuracy of the data presented herein. The contents do not necessarily reflect the official views of Toyota Motor North America.

The authors would like to sincerely thank Yongkang Liu and Ahmadreza Moradipari for their participation in the human-in-the-loop experiments and Hazem Abdelkawy for his support.

\bibliographystyle{IEEEtran}
\bibliography{references}
\end{document}